\documentclass[letterpaper]{article} % DO NOT CHANGE THIS
\usepackage{aaai24}  % DO NOT CHANGE THIS
\usepackage{times}  % DO NOT CHANGE THIS
\usepackage{helvet}  % DO NOT CHANGE THIS
\usepackage{courier}  % DO NOT CHANGE THIS
\usepackage[hyphens]{url}  % DO NOT CHANGE THIS
\usepackage{adjustbox}
\usepackage{graphicx} % DO NOT CHANGE THIS
\usepackage{natbib}  % DO NOT CHANGE THIS AND DO NOT ADD ANY OPTIONS TO IT
\usepackage{caption} % DO NOT CHANGE THIS AND DO NOT ADD ANY OPTIONS TO IT
\usepackage{algorithm,algpseudocode}
\usepackage{algorithmicx}
\usepackage{subcaption}
\usepackage{microtype}
\usepackage[dvipsnames]{xcolor}
\usepackage{amsmath,amsfonts}
\usepackage{newfloat}
\usepackage{listings}
\DeclareCaptionStyle{ruled}{labelfont=normalfont,labelsep=colon,strut=off} % DO NOT CHANGE THIS
\floatstyle{ruled}
\newfloat{listing}{tb}{lst}{}
\floatname{listing}{Listing}
\title{Minimum Coverage Sets for Training Robust Ad Hoc Teamwork Agents}
\author{
    Muhammad Rahman\textsuperscript{\rm 1},
    Jiaxun Cui\textsuperscript{\rm 1},
    Peter Stone\textsuperscript{\rm 1,2}
}
\affiliations{
    \textsuperscript{\rm 1}Department of Computer Science, The University of Texas at Austin\\
    \textsuperscript{\rm 2}Sony AI\\
    arrasy@cs.utexas.edu, cuijiaxun@utexas.edu, 
    pstone@cs.utexas.edu
}
\usepackage{algcompatible}
\usepackage{algorithm}
\usepackage{algpseudocode}

\usepackage{bibentry}
\begin{document}

\maketitle

\begin{abstract}
Robustly cooperating with unseen agents and human partners presents significant challenges due to the diverse cooperative conventions these partners may adopt. Existing Ad Hoc Teamwork (AHT) methods address this challenge by training an agent with a population of diverse teammate policies obtained through maximizing specific diversity metrics. However, prior heuristic-based diversity metrics do not always maximize the agent's robustness in all cooperative problems. In this work, we first propose that maximizing an AHT agent's robustness requires it to emulate policies in the minimum coverage set (MCS), the set of best-response policies to any partner policies in the environment. We then introduce the L-BRDiv algorithm that generates a set of teammate policies that, when used for AHT training, encourage agents to emulate policies from the MCS. L-BRDiv works by solving a constrained optimization problem to jointly train teammate policies for AHT training and approximating AHT agent policies that are members of the MCS. We empirically demonstrate that L-BRDiv produces more robust AHT agents than state-of-the-art methods in a broader range of two-player cooperative problems without the need for extensive hyperparameter tuning for its objectives.  Our study shows that L-BRDiv outperforms the baseline methods by prioritizing discovering distinct members of the MCS instead of repeatedly finding redundant policies.
\end{abstract}

\section{Introduction}
The \textit{Ad Hoc Teamwork}~(AHT) problem~\citep{stone2010adhoc} is concerned with learning ways to quickly cooperate with previously unseen agents or humans (henceforth referred to as ``\textit{unseen}'' or ``\textit{novel}'' teammates, or when unambiguous, simply “teammates”). In problems with multiple ways to coordinate, agents co-trained with a limited set of teammates may settle on cooperation conventions that only work when they collaborate with each other. Specialization towards these conventions diminishes an agent's ability to collaborate with unseen partners that adopt other conventions~\citep{hu2020otherplay}. 

Recent works address this problem by optimizing diversity metrics to generate sets of teammate policies for AHT training~\citep{lupu2021trajectory, strouse2021fititiouscoplay, xing2021entropy, bakhtin2022mastering}. Through interaction with the generated broadly representative teammate policies, an agent learns a policy to interact with previously unseen partners based on limited 
interactions. State-of-the-art methods optimize adversarial diversity to generate \emph{incompatible} teammate policies~\citep{charakorn2023generating,cui2023adversarial,rahman2023generating}. 
They seek sets of teammate policies, each maximizing their returns when playing with a designated AHT agent policy while minimizing returns with other policies.

Such existing diversity metrics are heuristic in nature and are not well-justified. It is unclear whether and how optimizing them can lead to improved robustness in general cooperative problems. We further demonstrate that optimizing these diversity metrics can fail to discover teammate policies under certain conventions even in simple cooperative games, specifically if following a convention yields high returns against the best-response policy to another generated teammate policy. Optimizing adversarial diversity can also generate teammates adopting \textit{self-sabotaging} policies~\citep{cui2023adversarial}. Self-sabotage potentially increases the difficulty of AHT training since the generated teammate policies can undermine collaboration with the trained AHT agent. 

In this work, we make three contributions that improve existing teammate generation methods for training robust AHT agents. First, we outline formal concepts describing an ideal set of teammate policies for training robust AHT agents, which can emulate the best-response policy to any teammate during interaction~\citep{chakraborty2014convergence}. The importance of finding the best-response policies to design a robust agent provides the motivation to estimate the \textbf{minimum coverage set (MCS)}, which is the set of best-response policies to any teammate policy in an environment, before interacting with unknown teammates. Second, we use the concept of MCS to propose the \textbf{L-BRDiv} algorithm\footnote{Implementation of L-BRDiv is available at \url{https://github.com/raharrasy/L-BRDiv}. The appendix is also accessible through \url{https://arxiv.org/abs/2308.09595}.} that jointly estimates the MCS of an environment and utilizes it to generate teammates for AHT training by solving a constrained optimization problem. L-BRDiv's generated set of teammate policies encourages AHT agents to emulate policies in the MCS through AHT training. Third, we provide experiments that empirically demonstrate that L-BRDiv produces more robust AHT agents than state-of-the-art teammate generation methods while requiring fewer hyperparameters to be tuned.

\section{Related Work}
\label{sec:RelWork}
\paragraph{Ad Hoc Teamwork} Assuming knowledge of teammate policies that will be encountered during evaluation, some existing AHT methods train adaptive AHT agents that can achieve near-optimal performance when interacting with any teammate policy encountered in evaluation~\citep{mirsky2022survey}. These methods equip an agent with two components. The first is a \textit{teammate modeling component} that infers an unknown teammate's policy via observations gathered from limited interactions with the unknown teammate. The second is an \textit{action selection component} that estimates the best-response policy to the inferred teammate policy, which selects actions that maximize the AHT agent's returns when collaborating with an unknown teammate. PLASTIC-Policy~\citep{AIJ16-Barrett} is an early example AHT method that defines an AHT agent policy based on the aforementioned components. Recent works~\citep{rahmanOpenAdHoc2021, zintgraf2021deep, papoudakis2021agent, gu2021online} implement these two components as neural network models which are trained to optimize the AHT agent's returns when dealing with a set of teammate policies seen during training. 

\paragraph{Adversarial Diversity}
Unlike the aforementioned AHT methods, our work assumes no knowledge of the potentially encountered teammate policies. Instead, our goal is to learn
what set of teammate policies,  when used in AHT training, maximizes the AHT
agent's robustness against previously unseen teammates.
Previous methods achieve this goal by optimizing \emph{Adversarial Diversity}~\citep{cui2023adversarial, charakorn2023generating, rahman2023generating}. Optimizing adversarial diversity maximizes \emph{self-play} returns, which are the expected returns when a generated policy $\pi^{-i}$ collaborates with its intended partner policy $\pi^{i}$. At the same time, adversarial diversity metrics also minimize \emph{cross-play} returns, the expected returns when $\pi^{-i}$ collaborates with the intended partner of another policy $\pi^{j}$. Creating teammate policies by optimizing adversarial diversity can be detrimental to AHT training for two reasons. First, minimizing cross-play returns can lead towards a self-sabotaging teammate policy, $\pi^{-i}$, that minimizes the returns when collaborating with anyone not behaving like its intended partner, $\pi^{-i}$. Learning to collaborate with a self-sabotaging $\pi^{-i}$ is difficult since learning to achieve high collaborative returns is only possible when the AHT agent fortuitously executes the same sequence of actions as $\pi^{i}$ during exploration. Second, we show in Section~\ref{sec:ExperimentAHTResults} and Appendix B that optimizing adversarial diversity will never yield teammate policies that lead towards the most robust AHT agent in certain environments.

\paragraph{Other Diversity-based Methods} Introducing diversity in training partners' policies is one way to generate robust response policies in multi-agent systems. A popular line of methods leverages population-based training and frequent checkpointing~\citep{strouse2021fititiouscoplay,vinyals2019grandmaster,cui2023macta, bakhtin2022mastering}. These methods rely on random seeds to find diverse policies, resulting in no guarantee that the generated policies are sufficiently diverse. Other studies optimize various types of diversity metrics directly into reinforcement learning objectives or as constraints. \citet{xing2021entropy} introduce a target-entropy regularization to Q-learning to generate information-theoretically different teammates. MAVEN~\citep{mahajan2019maven} maximizes the mutual information between the trajectories and latent variables to learn diverse policies for exploration. 
\citet{lupu2021trajectory} propose generating policies with different trajectory distributions. Trajectory diversity, however, is not necessarily meaningful for diversifying teammate policies~\citep{rahman2023generating}, so we do not consider these methods as baselines in our work. 

\section{Problem Formulation}
\label{sec:prob_form}
The interaction between agents in an AHT environment can be modeled as a decentralized partially observable Markov decision process (Dec-POMDP). A Dec-POMDP is defined by an 8-tuple, $\langle N,S,\{\mathcal{A}^{i}\}_{i=1}^{|N|}, P, R,\{\Omega^{i}\}_{i=1}^{|N|}, O, \gamma \rangle
$, with state space $S$, discount rate $\gamma$, and each agent $i\in{N}$ having an action space $\mathcal{A}^{i}$ and observation space $\Omega^{i}$. Each interaction episode between the AHT agent and its teammates starts at an initial state $s_{0}$ sampled from an initial state distribution $p_{0}(s)$. Denoting $\Delta(X)$ as the set of all probability distributions over set $X$, at each timestep $t$ agent $i$ cannot perceive $s_{t}$ and instead receives an observation $o^{i}_{t}\in\Omega^{i}$ sampled from the observation function, ${O: S\mapsto{\Delta(\Omega^{1}\times\dots\times\Omega^{|N|})}}$. Each agent $i\in{N}$ then decides its action at $t$, $a^{i}_{t}$, based on its policy, $\pi^{i}(H^{i}_{t})$, that is conditioned on the observation-action history of agent $i$, $H^{i}_{t} = \{o^{i}_{\leq{t}}, a^{i}_{<t}\}$. The action selected by each agent is then jointly executed as a joint action, $\mathbf{a}_{t}$. After executing $\mathbf{a}_{t}$, the environment state changes following the transition function, $P:S\times\mathcal{A}^{1}\times\dots\times\mathcal{A}^{|N|}\mapsto\Delta{S}$, and each agent receives a common scalar reward, $r_{t}$, according to the reward function, $R:S\times\mathcal{A}^{1}\times\dots\times\mathcal{A}^{|N|}\mapsto\mathbb{R}$.

Existing AHT methods learn policies for a robust AHT agent by interacting with teammate policies from the training teammate policy set, $\Pi^{\mathrm{train}} = \{\pi^{-1}, \pi^{-2},\dots,\pi^{-K}\}$. The AHT agent then optimizes its policy to maximize its returns in interactions with policies from $\Pi^{\mathrm{train}}$. The objective of these existing AHT methods can be formalized as:
\begin{equation} 
    \label{def:LearningObjPOAHT}
    \pi^{*,i}(\Pi^{\mathrm{train}}) =\underset{\pi^{i}}{\mathrm{argmax\ }}\mathbb{E}_{\substack{\pi^{-i}\sim{\mathbb{U}(\Pi^{\mathrm{train}})},\\ a^{i}_{t} \sim \pi^{i}, \\ a^{-i}_{t}\sim\boldsymbol{\pi}^{-i}, P, \, O}}\Bigg[\sum_{t=0}^{\infty} \gamma^{t}R(s_{t}, a_{t})\Bigg],
\end{equation}
with $\mathbb{U}(X)$ denoting a uniform distribution over set $X$. The learned AHT agent policy, $\pi^{*,i}(\Pi^{\mathrm{train}})$, is then evaluated for its robustness. Given an evaluated $\pi^{*,i}(\Pi^{\mathrm{train}})$, this robustness measure, $M_{\Pi^{\mathrm{eval}}}\left(\pi^{*,i}(\Pi^{\mathrm{train}})\right)$,  evaluates the expected returns when the AHT agent deals with teammates uniformly sampled from a previously unseen set of teammate policies, $\Pi^{\text{eval}}$. We formally define $M_{\Pi^{\mathrm{eval}}}\left(\pi^{*,i}(\Pi^{\mathrm{train}})\right)$ as the following expression:
\begin{equation}
    \label{def:robustness}
    \mathbb{E}_{\substack{\boldsymbol{\pi}^{-i}\sim{\mathbb{U}(\Pi^{\mathrm{eval}})}, a^{i}_{t} \sim \pi^{*,i}(\Pi^{\mathrm{train}}), \\a^{-i}_{t}\sim\boldsymbol{\pi}^{-i},\, P, O}}\Bigg[\sum_{t=0}^{\infty} \gamma^{t}R(s_{t}, a_{t})\Bigg],
\end{equation}
The dependence of $\pi^{*,i}(\Pi^{\mathrm{train}})$ on $\Pi^{\mathrm{train}}$ then implies that Expression~\ref{def:robustness} is also determined by $\Pi^{\mathrm{train}}$.

The goal of an AHT teammate generation process is to find $\Pi^{\mathrm{train}}$ producing an AHT agent policy that maximizes Expression~\ref{def:robustness} amid unknown $\Pi^{\text{eval}}$. Given the objective of AHT training from Equation~\ref{def:LearningObjPOAHT} and the definition of the robustness measure from Expression~\ref{def:robustness}, the objective of an AHT teammate generation process is to find the optimal set of training teammate policies, $\Pi^{*,\mathrm{train}}$, formalized as:
\begin{equation} \label{def:LearningObjPOAHTFinal}
    \underset{\Pi^{\text{train}}} {\mathrm{argmax}\ }\mathrm{ }\mathbb{E}_{\Pi^{\text{eval}}\sim \mathbb{U}(\Pi)}\left[M_{\Pi^{\mathrm{eval}}}\left(\pi^{*,i}(\Pi^{\mathrm{train}})\right)\right],
\end{equation}
While uniformly sampling $\Pi^{\text{train}}$ from $\Pi$ may appear to be a reasonable solution to produce $\Pi^{*,\mathrm{train}}$, training an AHT agent using $\Pi^{\text{train}}$ may produce low returns if we only sample a limited number of policies from $\Pi$. When $\Pi$ contains many possible teammate policies, the exact policies included in $\Pi^{\text{train}}$ becomes important to ensure that the AHT agent is robust when collaborating with any teammate policy in $\Pi$. 

\section{Creating Robust AHT Agents By Identifying Minimum Coverage Sets}
\label{sec:pipeline}
\begin{figure}[t]
\begin{subfigure}[ht]{0.38\linewidth}
\begin{center}
\includegraphics{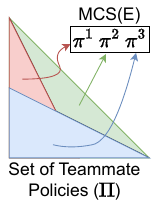} 
\subcaption{Best-response policies to each $\pi^{-i}\in\Pi$.}
\label{Fig:Clustering}
\end{center} 
\end{subfigure}
\hfill
\begin{subfigure}[ht]{0.57\linewidth}
\begin{center}
\includegraphics{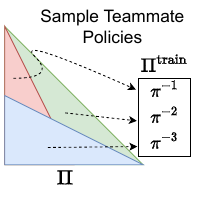} 
\subcaption{Generating $\Pi^{\text{train}}$ based on identified best-response policies.}
\label{Fig:Sampling}
\end{center}
\end{subfigure}

\begin{subfigure}[h]{1\linewidth}
\begin{center}
\includegraphics{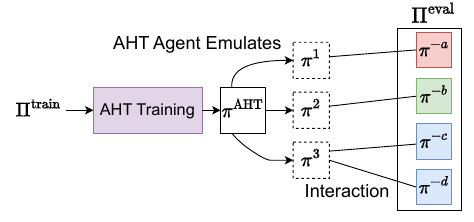} 
\subcaption{AHT training against $\Pi^{\text{train}}$ and the expected results when dealing with previously unseen teammate policies.}
\label{Fig:AHTTrain}
\end{center}
\end{subfigure}
\caption{Leveraging MCS(E) for Generating Robust AHT Agents. Figure~\ref{Fig:Clustering} visualizes how teammate policies (points in the large triangle) can be grouped based on their best-response policies. The rectangle then shows an example MCS(E). From each subset of $\Pi$ sharing the same best-response policy (colored small triangles), Figure~\ref{Fig:Sampling} visualize how one policy is sampled from each subset to create $\Pi^{\text{train}}$ for AHT training. As visualized in Figure~\ref{Fig:AHTTrain}, using our generated $\Pi^{\text{train}}$ for AHT training should encourage agents that emulate the best-response policy (dashed squares) to any $\pi^{-i}\in\Pi$ when dealing teammates from $\Pi^{\text{eval}}$ (squares whose color represent its best-response policy).} 
\label{Figs:MCSUse}
\end{figure}
Assuming knowledge of $\Pi^{\text{eval}}$, the robustness of an AHT agent as defined by Expression~\ref{def:robustness} can be optimized by using $\Pi^{\text{eval}}$ as teammate policies for AHT training. Given a teammate modeling component that accurately infers an unknown teammate's policy from $\Pi^{\text{eval}}$ and an action selection component that can emulate any policy in the set of best-response policies to policies in $\Pi^{\text{eval}}$, $\text{BR}(\Pi^{\text{eval}})$, an AHT agent's robustness is maximized by following the best-response policy to the inferred teammate policy. Unfortunately, $\Pi^{\text{eval}}$ being unknown makes this ideal training process impossible.

Improving an AHT agent's robustness without knowing $\Pi^{\text{eval}}$ is still possible by identifying the \textit{coverage set} of an environment. Denoting an environment characterized by a Dec-POMDP as $\text{E}$, any set containing at least one best-response policy to each teammate policy in $\Pi$ is a coverage set of an environment, CS(E). CS(E) is formally characterized as:
\begin{equation}
\begin{split}
    \label{def:coverage_set}
     &\forall{\pi^{-i}}\in \Pi, \forall{H_{t}}, \exists{\pi^{*}} \in \text{CS(E)}: \\ &\mathbb{E}_{\substack{s_{0}\sim{p_{0}}}}\left[\textbf{R}_{{*}, {-i}}(H_{t})\right] = \underset{\pi^{i}\in\Pi}{\text{max} } \mathbb{E}_{\substack{s_{0}\sim{p_{0}}}}\left[\textbf{R}_{, {i}, {-i}}(H_{t})\right],
\end{split}
\end{equation}
where $\textbf{R}_{{i}, {-i}}(H)$ denotes the following expression:
\begin{equation}\mathbb{E}_{\substack{a^{i}_{T}\sim\pi^{i}(.|H_{T}),\\ a^{-i}_{T} \sim\pi^{-i}(.|H_{T}), \\ P, O}} \left[\sum_{T=t}^{\infty}\gamma^{T-t}R_{T}(s_{T}, a_{T})\middle| H_{t}=H\right].
\end{equation} 
Given this definition, a CS(E) remains a coverage set when policies are added. Thus, $\Pi$ itself is trivially a coverage set.

Irrespective of $\Pi^{\text{eval}}$, CS(E) will contain at least a single best-response policy to any $\pi^{-i}\in\Pi^{\text{eval}}$ since $\Pi^{\text{eval}}\subseteq\Pi$. An AHT agent capable of emulating any policy from CS(E) consequently can follow any policy from $\text{BR}(\Pi^{\text{eval}})$ for any $\Pi^{\text{eval}}$. Therefore, training an AHT agent to emulate any policy from CS(E) gives us a solution to design robust AHT agents even when $\Pi^{\text{eval}}$ is unknown.

Considering CS(E) may contain policies that are not a best-response policy to any member of $\Pi$, we ideally only train AHT agents to emulate a subset of CS(E) that consists of policies that are the best-response to some $\pi^{-i}\in\Pi$. Based on this idea, we define the \textit{minimum coverage set} of an environment, $\text{MCS(E)}\subseteq \Pi$, that is a coverage set ceasing to be a coverage set if any of its elements are removed. This characteristic of $\text{MCS(E)}$ is formalized as:
\begin{equation}
    \label{def:mcs}
    \forall{\pi^{i}\in{\text{MCS(E)}}}: \text{MCS(E)} - \{\pi^{i}\} \text{ is not a coverage set}.
\end{equation}
In the example provided in Figure~\ref{Fig:Clustering}, $\text{MCS(E)}=\{\pi^{1}, \pi^{2}, \pi^{3}\}$ is an MCS since the elimination of any policy, $\pi$, from it cause a subset of $\Pi$ to not have their best-response policy in $\text{MCS(E)}-\{\pi\}$.

Our work aims to design AHT agents capable of emulating any policies from MCS(E) by constructing $\Pi^{\text{train}}$ in a specific way. If $\Pi^{\text{train}}$ is constructed for each $\pi^{i}\in\text{MCS(E)}$ to have a $\pi^{-i}\in\Pi^{\text{train}}$ such that $\pi^{i}\in\text{BR}(\{\pi^{-i}\})$, using $\Pi^{\text{train}}$ while optimizing Equation~\ref{def:LearningObjPOAHT} enables us to achieve this goal. The role of MCS(E) in our teammate generation process is visualized in Figures~\ref{Fig:Sampling} and~\ref{Fig:AHTTrain}. 

\section{L-BRDiv: Generating Teammate Policies By Approximating Minimum Coverage Sets}
\begin{figure*}[ht]
    \centering
    \includegraphics{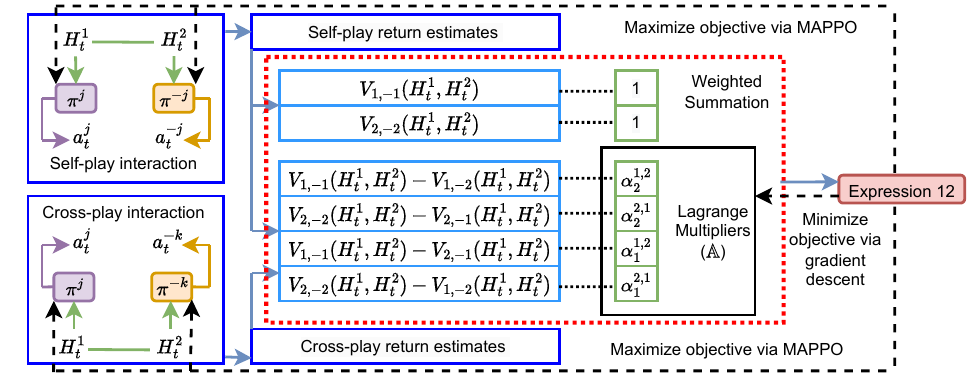}
    \caption{Lagrangian Best Response Diversity (L-BRDiv). The L-BRDiv algorithm trains a collection of policy networks (purple and orange boxes) and Lagrange multipliers (green cells inside the black rectangle). The purple boxes represent a policy from $\{\pi^{i}\}_{i=1}^{K}\subseteq\Pi$ while the policies visualized as an orange box is from $\{\pi^{-i}\}_{i=1}^{K}\subseteq\Pi$. Estimated returns between any possible pairs of policy, $(\pi^{j},\pi^{-k}) \in (\{\pi^{i}|\pi^{i}\in\Pi\}_{i=1}^{K} \times \{\pi^{-i}|\pi^{-i}\in\Pi\}_{i=1}^{K})$, and their associated Lagrange multipliers are used to compute the optimized term in the Lagrangian dual form (right red box) via a weighted summation operation (black dotted lines connect weights and multiplied terms). The policy networks are then trained via MAPPO~\citep{yu2022the} to maximize this optimized term, while the Lagrange multipliers are trained to minimize the term via stochastic gradient descent.}
    \label{fig:method_desc}
\end{figure*}
        
This section introduces our proposed teammate generation method based on estimating MCS(E). Section~\ref{sec:mcs_estimate} details a constrained objective we use to estimate MCS(E). Finally, Section~\ref{sec:team_gen_mcs} provides a method that solves the constrained objective to jointly estimate MCS(E) while generating $\Pi^{\text{train}}$. 

\subsection{Jointly Approximating MCS(E) and Generating $\Pi^{\text{train}}$}
\label{sec:mcs_estimate}

Discovering $\text{MCS(E)}$ by enumerating the AHT agent's best-response policy to each teammate policy is intractable given the infinite policies in $\Pi$. Instead, we can estimate $\text{MCS(E)}$ by eliminating policies from a finite $\text{CS(E)}$ to generate $\text{MCS(E)}$. %Given a finite $\text{CS(E)}$, we can remove an AHT agent policy from being considered a member of $\text{MCS(E)}$ if is not the best response to any teammate policy. 
Given a finite $\text{CS(E)}$, an AHT agent policy is not a member of $\text{MCS(E)}$ if it is not the best response to any teammate policy. 

We check if $\pi^{i}\in\text{CS(E)}$ is the best-response policy of at least one policy from $\Pi$ by solving the \textit{feasibility problem}, which is the following constrained optimization problem: 
\begin{equation}
    \label{def:indiv_lagrange}
    \underset{\pi^{-i}\in\Pi}{\text{max}} \mathbb{E}_{s_{0}\sim{p_{0}}} \left[\textbf{R}_{{i},{-i}}(H_{t})\right], 
\end{equation}
with the following constraints:
\begin{equation}
 \label{def:indiv_lagrange_const}
 \begin{split}
 &\forall{\pi^{j}\in(\text{CS(E)}-\{\pi^{i}\}}): \\ &\mathbb{E}_{\substack{s_{0}\sim{p_{0}}}}\left[\textbf{R}_{{j},{-i}}(H_{t})\right] \leq \mathbb{E}_{s_{0}\sim{p_{0}}} \left[\textbf{R}_{{i},{-i}}(H_{t})\right].
 \end{split}
\end{equation}
Any CS(E) member that violates the above constraint for all $\pi^{-i}\in\Pi$ is not a member of MCS(E). While this approach relies on knowing a finite $\text{CS(E)}$, note that knowledge of a finite $\text{CS(E)}$ is sometimes available. For instance, the set of all deterministic policies is a finite $\text{CS(E)}$ for environments with a finite action space and state space. 

Applying the above procedure to find MCS(E) can still be impossible for two reasons. First, a finite $\text{CS(E)}$ can be unknown. Second, the size of $\text{CS(E)}$ may be prohibitively large, which prevents solving the feasibility problem for all $\pi^{i}\in\text{CS(E)}$. Amid these challenging problems, we resort to estimating $\text{MCS(E)}$ by only discovering its subset with $K$ policies, $\text{MCS}^{\text{est}}\text{(E)} = \{\pi^{i}\}_{i=1}^{K}$.

We now describe an alternative constrained optimization objective that jointly finds $\text{MCS}^{\text{est}}\text{(E)}$ while generating a set of teammate policies for AHT training, $\Pi^{\text{train}} = \{\pi^{-i}\}_{i=1}^{K}$, according to the method illustrated in Figure~\ref{Figs:MCSUse}. Two characteristics are desired when finding $\text{MCS}^{\text{est}}\text{(E)}$. First, we require each AHT agent policy from $\text{MCS}^{\text{est}}\text{(E)}$ to only be the best-response policy to one teammate policy from $\Pi^{\text{train}}$, $\pi^{i}$. The second characteristic prioritizes the discovery of $\text{MCS(E)}$ members that enables the AHT agent to produce high returns with a designated teammate policy, $\pi^{-i}\in\Pi$. These two requirements are formulated as the following constrained optimization problem: 
\begin{equation}
\label{Eq:obj}
\underset{\substack{\{\pi^{i}\}_{i=1}^{K}\subseteq\Pi,\\ \{\pi^{-i}\}_{i=1}^{K}\subseteq\Pi}}{\text{max}} \sum_{i\in\{1,2,\dots,K\}}\mathbb{E}_{s\sim{p_{0}}}\left[\textbf{R}_{{i}, {-i}}(H_{t})\right],
\end{equation}
with the following constraints that must be fulfilled for all $i,j\in\{1,2,\dots,K\}$ and $i\neq{j}$:
\begin{equation}
\label{Eq:Constraints1}
\mathbb{E}_{s\sim{p_{0}}}\left[\textbf{R}_{{j},{-i}}(H_{t})\right] + \tau \leq \mathbb{E}_{s\sim{p_{0}}}\left[\textbf{R}_{{i},{-i}}(H_{t})\right],
\end{equation}
\begin{equation}
\label{Eq:Constraints2}
\mathbb{E}_{s\sim{p_{0}}}\left[\textbf{R}_{{i},{-j}}(H_{t})\right] + \tau \leq \mathbb{E}_{s\sim{p_{0}}}\left[\textbf{R}_{{i},{-i}}(H_{t})\right].
\end{equation}
Note that a near-zero positive threshold  ($\tau > 0$) is introduced in the constraints to prevent discovering duplicates of the same $\pi^{i}$ and $\pi^{-i}$, which turns Constraints~\ref{Eq:Constraints1} \&~\ref{Eq:Constraints2} into equality when $\tau = 0$.

\subsection{Lagrangian BRDiv (L-BRDiv)}
\label{sec:team_gen_mcs}

We present the \textbf{L}agrangian \textbf{B}est \textbf{R}esponse \textbf{Div}ersity (L-BRDiv) algorithm to generate $\Pi^{\text{train}}$ that encourages an AHT agent to emulate $\text{MCS}^{\text{est}}\text{(E)}$. L-BRDiv generates $\Pi^{\text{train}}$ by solving the Lagrange dual of the optimization problem specified by Expressions~\ref{Eq:obj}-\ref{Eq:Constraints2}, which is an unconstrained objective with the same optimal solution. 
The Lagrange dual for our optimization problem is defined as:
\begin{align}
    \label{def:dual_lagrange}
    &\underset{\substack{\mathbb{A}\subseteq\mathbb{R}_{\geq 0}^{K(K-1)}\\\times\mathbb{R}_{\geq 0}^{K(K-1)}}}{\text{min}}\underset{\substack{\{\pi^{i}\}_{i=1}^{K}\subseteq{\Pi},\\ \{\pi^{-i}\}_{i=1}^{K}\subseteq{\Pi}}}{\text{max}} \Big( \sum_{i\in\{1,\dots,K\}} \mathbb{E}_{s_{0}\sim{p_{0}}}\left[\textbf{R}_{{i},{-i}}(H_{t})\right] + \nonumber\\ &\sum_{\substack{i,j\in\{1,\dots,K\}\\i\neq j}} \text{ }\alpha_{1}^{i,j} \left(  \mathbb{E}_{s_{0}\sim{p_{0}}}\left[\textbf{R}_{i,-i}(H_{t}) - \tau - \textbf{R}_{{j},{-i}}(H_{t})\right] \right) + \nonumber\\
    &\sum_{\substack{i,j\in\{1,\dots,K\}\\i\neq j}} \text{ }\alpha_{2}^{i,j}\left(\mathbb{E}_{s_{0}\sim{p_{0}}}\left[\textbf{R}_{{i},{-i}}(H_{t}) - \tau - \textbf{R}_{{i},{-j}}(H_{t})\right]\right)\Big),
\end{align}
with $\mathbb{A} = \{(\alpha_{1}^{i,j}, \alpha_{2}^{i,j})| \alpha_{1}^{i,j}\geq{0}, \alpha_{2}^{i,j}\geq{0}\}_{i,j\in\{1,2,\dots,K\}, i\neq{j}}$ denoting the set of optimizable Lagrange multipliers.

L-BRDiv learns to assign different values to Lagrange multipliers in $\mathbb{A}$ of (\ref{def:dual_lagrange}). Optimizing Lagrange multipliers gives L-BRDiv two advantages over previous methods, which treat these hyperparameters as constants. First, we demonstrate in Section~\ref{sec:ExperimentDetails} that L-BRDiv creates better $\Pi^{\text{train}}$ by identifying more members of MCS(E). Second, it does not require hyperparameter tuning on appropriate weights associated with cross-play return, which in previous methods require careful tuning to discover members of MCS(E)~\citep{rahman2023generating} and prevent the generation of incompetent policies not achieving high returns against any AHT agent policy~\citep{charakorn2023generating}.

We detail L-BRDiv's teammate generation process in Algorithm~\ref{alg:A2COpt} and analyze its computational complexity in Appendix D. L-BRDiv implements the policies optimized in the Lagrange dual as neural networks trained with MAPPO~\citep{yu2022the} to maximize the weighted advantage function (\ref{Eq:WeightedAdvantage}), whose weights correspond to the total weight associated with each expected return term in (\ref{def:dual_lagrange}). At the same time, L-BRDiv trains a critic network to bootstrap the evaluation of (\ref{def:dual_lagrange}) instead of a Monte Carlo approach, which can be expensive since it requires all generated policy pairs to initially follow the observation-action history, $H_{t}$. Meanwhile, the Lagrange multipliers are trained to minimize (\ref{def:dual_lagrange}) while ensuring it is non-negative. Figure~\ref{fig:method_desc} then summarizes the training process of L-BRDiv's models.

\begin{algorithm}[t]
\caption{Lagrangian Best Response Diversity}\label{alg:A2COpt}
\begin{algorithmic}[1]
\Require 
\Statex Cardinality of $\text{MCS}^{\text{est}}(\text{E})$ and $\Pi^{\text{train}}$, $K$. 
\Statex Randomly initialized policy networks in $\text{MCS}^{\text{est}}(\text{E})$ \& $\Pi^{\text{train}}$, denoted by $\{\pi^{i}_{\theta_{i}}\}_{i=1}^{K}$ \& $\{\pi^{-i}_{\theta_{-i}}\}_{i=1}^{K}$ respectively.
\Statex Randomly initialized critic network $V^{j,-i}_{\theta_{c}}$, target $V^{j,-i}_{\theta'_{c}}$.
\Statex Initial values for the Lagrange multipliers, $\mathbb{A}$.
\FOR{$t_{\text{update}}=1,2,\dots,N_{\text{updates}}$}
\STATE $(i,j) \sim \mathbb{U}(\{1,2,\dots,K\}^{2})$
\STATE $D \leftarrow \text{AgentInteraction}(\pi^{j}_{\theta_{j}}, \pi^{-i}_{\theta_{-i}})$
\FOR{$(H_{t}, a_{t}, r_{t}, H_{t+1})\in D$}
\State {// Critic \& Policy Optimization Step (Lines 6 \& 8)}
\State Update $\theta_{c}$ with SGD \& a target critic to minimize
\begin{equation}
\left(V^{j,-i}_{\theta_{c}}(H_{t}) - r_{t} -\gamma V^{j,-i}_{\theta'_{c}}(H_{t+1})\right)^{2}    
\end{equation}
\State $
w^{i,j}(\mathbb{A})\leftarrow \begin{cases}
  1+\underset{k\neq{j}}{\sum}\left(\alpha^{i,k}_{1}+\alpha^{i,k}_{2}\right)&, i=j \\
  -\left(\alpha^{i,j}_{1}+\alpha^{j,i}_{2}\right) &, i\neq{j}
\end{cases}$
\State Update $\theta_{j}$ and $\theta_{-j}$ with MAPPO to maximize:
\begin{equation}
\label{Eq:WeightedAdvantage}
    w^{i,j}(\mathbb{A})\left(r_{t}+\gamma V_{\theta_{c}}^{j,-i}(H_{t+1}) - V^{j,-i}_{\theta_{c}}(H_{t})\right)
\end{equation}
\IF{$t_{\text{update}}$ mod $T_{\text{lagrange}}$ = 0}
\State {{// Lagrange Multiplier Optimization Step}}
\State Update $\mathbb{A}$ using SGD to minimize Expression~\ref{def:dual_lagrange} where $\forall i,j\in\{1,2,\dots,K\}$:
\begin{equation}
    \mathbb{E}_{s_{0}\sim{p_{0}}}\left[\textbf{R}_{{j},{-i}}(H_{t})\right] \approx V^{j,-i}_{\theta_{c}}(H_{t})
\end{equation}
\State $\mathbb{A} \leftarrow \{\text{max}(\alpha,0)\ |\ \alpha\in\mathbb{A}\}$
\ENDIF
\ENDFOR
\ENDFOR
\State \textbf{Return} $\{\pi^{-i}_{\theta_{-i}}\}_{i=1}^{K}$
\end{algorithmic}
\end{algorithm}

\section{Experiments}
\label{sec:ExperimentDetails}
In this section, we describe the environments and baseline algorithms in Sections~\ref{sec:ExperimentEnvironments} and~\ref{sec:ExperimentBaselines}. Section~\ref{sec:ExperimentSetup} then details the experiment setups for evaluating the robustness of AHT agents in L-BRDiv and baseline methods via their generated training teammate policies. Finally, we present the AHT experiment results and an analysis of MCS${}^{\text{est}}$(E) policies identified by L-BRDiv in Sections~\ref{sec:ExperimentAHTResults} and~\ref{sec:ExperimentBehaviourAnalysis}.

\subsection{Environments}
\label{sec:ExperimentEnvironments}
\begin{figure}[ht]
\begin{subfigure}[t]{0.64\linewidth}
\begin{center}
\begin{adjustbox}{width=0.5\textwidth}
\begin{tabular}{|c|c|c|}
\hline
10 & 0 & 4 \\ \hline
0  & 6 & 4 \\ \hline
4  & 4 & 6 \\ \hline
\end{tabular}
\end{adjustbox}
\vspace{0.05cm}
\subcaption{Repeated Matrix Game.}
\label{Fig:RPMEnv}
\end{center} 
\end{subfigure}
\hfill
\begin{subfigure}[ht]{0.3\linewidth}
\begin{center}
\includegraphics[width=0.9\linewidth]{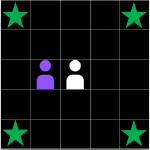} 
\subcaption{Coop Reaching.}
\label{Fig:CoopReachingEnv}
\end{center}
\end{subfigure}

\begin{subfigure}[b]{0.64\linewidth}
\begin{center}
\begin{adjustbox}{width=0.83\linewidth}
\begin{tabular}{ccccc}
\multicolumn{1}{l}{}   & A                       & B                       & C                      & D                      \\ \cline{2-5} 
\multicolumn{1}{c|}{A} & \multicolumn{1}{c|}{10} & \multicolumn{1}{c|}{0}  & \multicolumn{1}{c|}{6} & \multicolumn{1}{c|}{6} \\ \cline{2-5} 
\multicolumn{1}{c|}{B} & \multicolumn{1}{c|}{0}  & \multicolumn{1}{c|}{10} & \multicolumn{1}{c|}{6} & \multicolumn{1}{c|}{6} \\ \cline{2-5} 
\multicolumn{1}{c|}{C} & \multicolumn{1}{c|}{6}  & \multicolumn{1}{c|}{6}  & \multicolumn{1}{c|}{8} & \multicolumn{1}{c|}{0} \\ \cline{2-5} 
\multicolumn{1}{c|}{D} & \multicolumn{1}{c|}{6}  & \multicolumn{1}{c|}{6}  & \multicolumn{1}{c|}{0} & \multicolumn{1}{c|}{8} \\ \cline{2-5} 
\end{tabular}
\end{adjustbox}
\subcaption{Weighted Coop Reaching.}
\label{Fig:WeightedCoopReachingEnv}
\end{center}
\end{subfigure}
\hfill
\begin{subfigure}[b]{0.3\linewidth}
\begin{center}
\includegraphics[width=0.9\textwidth]{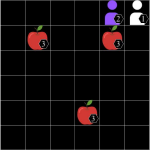} 
\subcaption{LBF.}
\label{Fig:LBFEnv}
\end{center}
\end{subfigure}
\caption{Environments Used in AHT Experiments. We provide experiments in a repeated matrix game whose reward function is displayed in Figure~\ref{Fig:RPMEnv}. Figure~\ref{Fig:CoopReachingEnv} displays an example state of the Cooperative Reaching environment where the green stars represent corner cells that provide agents rewards once they simultaneously reach it. If we start from the top-left corner cell in Figure~\ref{Fig:CoopReachingEnv} and assign IDs (A-D) to corner cells in a clockwise manner, Figure~\ref{Fig:WeightedCoopReachingEnv} shows the reward function of the Weighted Cooperative Reaching environment where agents' rewards depend on which pair of destination cells the two agents arrive at. Finally, Figure~\ref{Fig:LBFEnv} shows a sample state of Level-based Foraging (LBF) where the apples represent the collected objects.}
\label{Figs:EnvFigs}
\end{figure}
We run our experiments in four two-player cooperative environments. The first environment is a repeated matrix game where agents have three actions, whose reward function is provided in Figure~\ref{Fig:RPMEnv}. Since eliminating self-sabotaging behaviour~\citep{cui2023adversarial} is not the focus of our work, we remove teammate-related information and actions from an agent's observation such that self-sabotaging behaviour is not a member of possibly discovered teammate behaviours, $\Pi$. We also do experiments in the Cooperative Reaching environment~\citep{rahman2023generating} where two agents can move across the four cardinal directions in a two-dimensional grid world. Both agents are given a reward of 1 once they simultaneously arrive at the same corner grid. The third environment is Weighted Cooperative Reaching, which is similar to Cooperative Reaching except for a modified reward function (Figure~\ref{Fig:WeightedCoopReachingEnv}) that provides lower rewards if both agents arrive at different corner cells. The last environment is Level-based Foraging (LBF)~\citep{christianos2020shared}, where both agents must move along the four cardinal directions to a cell next to the same object and retrieve it by simultaneously selecting actions for collecting objects. Successful object collection gives both agents a reward of 0.33.

\subsection{Baseline Methods}
\label{sec:ExperimentBaselines}

Our experiments compare L-BRDiv against methods that maximize adversarial diversity, such as BRDiv~\citep{rahman2023generating} and LIPO~\citep{charakorn2023generating}. 
Comparing L-BRDiv and BRDiv helps investigate the detrimental effect of using fixed uniform weights instead of L-BRDiv's optimized Lagrange multipliers ($\mathbb{A}$). Meanwhile, including LIPO as a baseline enables us to investigate the advantage of L-BRDiv and BRDiv's use of weights with a larger magnitude for self-play maximization (i.e. $w^{i,i}(\mathbb{A})$ in Eq.~\ref{Eq:WeightedAdvantage}) compared to the weights for cross-play minimization (i.e. $w^{i,j}(\mathbb{A})$ in Eq.~\ref{Eq:WeightedAdvantage}). We do not compare our method with ADVERSITY~\citep{cui2023adversarial}, which combines LIPO with techniques to prevent self-sabotage. We hold that self-sabotaging policies should not be ruled out during policy generation since teammates may still use them. By not preventing the discovery of such policies, we ensure that our method remains fully general.

\subsection{Experiment Setup}
\label{sec:ExperimentSetup}
We start our experiments for each environment by generating $K$ training teammate policies using the compared methods. We ensure fairness in our experiments by using $\text{RL}^{2}$ algorithm~\citep{duan2016rl} to find an optimal AHT agent policy defined in Equation~\ref{def:LearningObjPOAHT} based on $\Pi^{\text{train}}$ generated by each teammate generation algorithm. Since our partially observable environments provide no useful information to infer teammate policies except for rewards obtained at the end of each interaction episode, we choose $\text{RL}^{2}$ since it can use reward information to create agent representations maintained and updated across multiple episodes. For each of the compared algorithms, the teammate generation and AHT training process are repeated under four seeds to allow for a statistically sound comparison between each method's performance. As a measure of robustness, we then evaluate the average returns of the AHT agent trained from each experiment seed when collaborating with policies sampled from $\Pi^{\text{eval}}$. We construct $\Pi^{\text{eval}}$ for each environment by creating heuristic-based agents, whose behaviour we describe in Appendix A. Finally, we compute the mean and 95\% confidence interval of the recorded returns across four seeds and report it in Figure~\ref{Figs:Results}.

\subsection{Ad Hoc Teamwork Experiment Results}
\label{sec:ExperimentAHTResults}

\begin{figure*}[ht]
\includegraphics{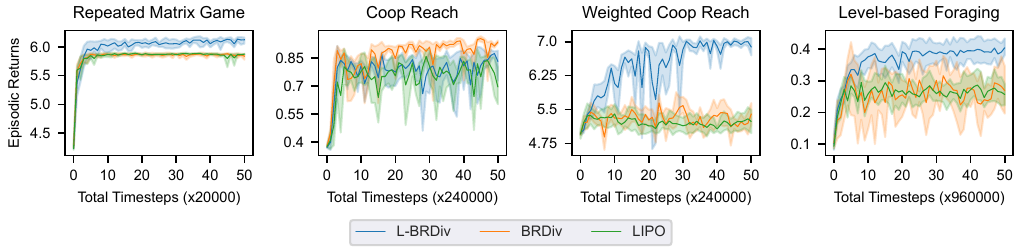} 

\caption{Generalization Performance Against Previously Unseen Teammate Types. This figure shows that L-BRDiv produced significantly higher episodic returns when dealing with unknown teammate policies in all environments except for Cooperative Reaching. We also show L-BRDiv achieving similar returns to other methods in Cooperative Reaching.} 
\label{Figs:Results}
\end{figure*}

\begin{figure}[t]
\begin{subfigure}[h]{\linewidth}
\begin{center}
\begin{adjustbox}{width=0.5\textwidth}
\begin{tabular}{|c|c|c|c|}
\hline
  & $\pi$(A)  & $\pi$(B)  & $\pi$(C)  \\ \hline
1 & 1 & 0  & 0  \\ \hline
2 & 0  & 1 & 0  \\ \hline
3 & 0  & 0  & 1 \\ \hline
\end{tabular}
\end{adjustbox}
\subcaption{AHT agent action selection probability for policies in MCS${}^{\text{est}}$(E)  in the Repeated Matrix Game.}
\label{Fig:MatrixActionFrequency}
\end{center} 
\end{subfigure}
\begin{subfigure}[h]{0.48\linewidth}
\begin{center}
\includegraphics[width=0.60\linewidth]{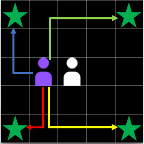} 
\subcaption{MCS${}^{\text{est}}$(E) in Coop Reaching \& Weighted Coop Reaching.}
\label{Fig:CoopReachingMCS}
\end{center}
\end{subfigure}
\begin{subfigure}[h]{0.48\linewidth}
\begin{center}
\includegraphics[width=0.60\linewidth]{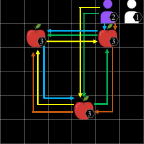} 
\subcaption{AHT agent policies in the MCS${}^{\text{est}}$(E) discovered for LBF.}
\label{Fig:LBFUnique}
\end{center}
\end{subfigure}  

\caption{MCS${}^{\text{est}}$(E) Yielded by L-BRDiv. L-BRDiv is capable of estimating all members of MCS(E) in all environments except LBF. Even so, L-BRDiv still discovers more conventions with distinct best-response policies than the baselines in LBF. The discovery of more MCS(E) results in L-BRDiv producing more robust AHT agents.}
 
\end{figure}

Figure~\ref{Figs:Results} shows the results of the AHT experiments. We find that L-BRDiv significantly outperforms other compared methods in the repeated matrix game, Weighted Cooperative Reaching, and LBF. While BRDiv slightly outperforms L-BRDiv in Cooperative Reaching, overlapping confidence intervals among the last few checkpoints suggest that the difference is only marginally significant.

L-BRDiv outperforms the compared baselines in all environments except Cooperative Reaching since these environments all have reward functions that cause some members of the MCS, $\pi^{i}\in\text{MCS(E)}$, to yield high expected returns in cross-play interactions against a generated teammate policy, $\pi^{-j}\in\Pi^{\text{train}}$, that is not its intended partner, $\pi^{-i}\in\Pi^{\text{train}}$. Meanwhile, all $\pi^{i}\in$MCS(E) for Cooperative Reaching have equally low (i.e. zero) returns against the intended partner of other MCS(E) members. The large cross-play returns disincentivize BRDiv and LIPO's optimized objective from discovering $\pi^{i}$ and $\pi^{-i}$ during teammate generation. The inability to discover $\pi^{i}\in\text{MCS(E)}$ and $\pi^{-i}$ will then lead towards diminished robustness since the trained AHT agent will yield lower returns against teammates whose best-response policy is $\pi^{i}$. In contrast, Cooperative Reaching's reward structure makes MCS(E) (i.e. the set of four policies moving towards each distinct corner cell) consist of policies yielding equally low cross-play returns of zero among each other.

Although both BRDiv and LIPO are equipped with a hyperparameter, $\alpha>0$, that can change weights associated with self-play returns maximization and cross-play returns minimization in their learning objective, it is possible to find simple scenarios where no feasible $\alpha$ facilitates the discovery of a desirable $\Pi^{\text{train}}$ to maximize an AHT agent's robustness. Such a desirable $\Pi^{\text{train}}$ is characterized by all AHT agent policies in MCS(E) having at least one teammate policy in $\in\Pi^{\text{train}}$ whom it is the best-response policy to. Appendix B shows that the Repeated Matrix Game and Weighted Cooperative Reaching environment are examples of such scenarios. Even in environments like LBF where there may exist an $\alpha$ enabling both BRDiv and LIPO to discover a desirable $\Pi^{\text{train}}$ by optimizing their learning objectives, finding an appropriate $\alpha$ is costly if we factor in the computational resources required to run a single teammate generation process. Unlike BRDiv and LIPO, L-BRDiv's inclusion of Lagrange multipliers as learned parameters enables it to discover desirable $\Pi^{\text{train}}$ in a wider range of environments while reducing the number of hyperparameters that must be tuned.

Note that L-BRDiv and the baseline methods all successfully discover MCS(E) in Cooperative Reaching. However, each teammate policy generated by L-BRDiv and LIPO which has one of the MCS(E) members as its best-response policy ends up being less optimal than their BRDiv-generated counterparts. These suboptimal policies require more steps to complete an episode by occasionally moving away from their destination corner cell. Learning from these suboptimal agents made the AHT agent less decisive when selecting which corner cell to move towards and finally ends up producing agents with slightly lower returns.

\subsection{Behaviour Analysis}
\label{sec:ExperimentBehaviourAnalysis}

The AHT agent policies that L-BRDiv discovers as members of MCS${}^\text{est}$ in all environments are provided in Figures~\ref{Fig:MatrixActionFrequency}-\ref{Fig:LBFUnique}. Unlike the compared baseline methods that only discover two members of MCS(E), results from the Repeated Matrix Game show L-BRDiv is capable of consistently finding all three deterministic policies that are members of MCS(E). As a consequence of Cooperative Reaching's reward structure, all compared methods successfully discover MCS(E) and achieve the similar performances. Meanwhile, L-BRDiv is the only method that finds all four members of MCS(E) corresponding to movement towards each corner grid in Weighted Cooperative Reaching. As we show in Appendix B, BRDiv and LIPO's failure to discover all members of MCS(E) in the Repeated Matrix Game and Weighted Cooperative Reaching is because discovering MCS(E) does not optimize their optimized objective for any constant and uniform $\alpha$. Despite no method perfectly discovering MCS(E) consisting of all six possible orderings for collecting objects in LBF, L-BRDiv is closer to estimating MCS(E) than the baseline algorithms by discovering four MCS(E) members in one seed and five MCS(E) members in the remaining seeds. L-BRDiv's ability to discover more MCS(E) members than baselines leads towards more robust AHT agents that can emulate the best-response policy to a wider range of teammate policies.

\section{Conclusion \& Future Work}
%This work introduces L-BRDiv, a method that generates teammate policies to train robust AHT agents. 
In this work, we propose that an appropriate set of teammate policies for AHT training must enable agents to emulate all policies in MCS(E), the smallest set of policies containing the best-response policy to any teammate policy in $\Pi$. 
To generate such teammate policies for robust AHT training, we introduce and evaluate L-BRDiv. By solving a constrained optimization problem using the Lagrange multiplier technique, L-BRDiv then learns to jointly approximate the MCS of an environment and generate a set of teammate policies for AHT training. Our experiments indicate that L-BRDiv yields more robust AHT agents compared to state-of-the-art teammate generation methods by identifying more members of the MCS while also removing the need for tuning important hyperparameters used in prior methods.

Future work will consider extending L-BRDiv to more complex environments where more than two agents must collaborate. Another promising research direction is to extend L-BRDiv with techniques to discourage the discovery of self-sabotaging policies~\citep{cui2023adversarial}. Finally, applying our method in fully competitive and general-sum games is another promising direction for creating robust agents since the concept of minimum coverage sets is not limited to fully cooperative problems.

\section*{Acknowledgements}
All research conducted in this work was done under the Learning Agents Research Group (LARG) at UT Austin's Department of Computer Science. The research in LARG is supported in part by Lockheed Martin, NSF (CPS-1739964, IIS-1724157, NRI-1925082), ONR (N00014-18-2243), FLI (RFP2-000), ARO (W911NF19-2-0333), DARPA, GM, and Bosch. Peter Stone serves as the Executive Director of Sony AI America and receives financial compensation for this work. The terms of this arrangement have been reviewed and approved by the University of Texas at Austin following its policy on objectivity in research.

\bibliography{aaai24}

\begin{thebibliography}{22}
\providecommand{\natexlab}[1]{#1}

\bibitem[{Bakhtin et~al.(2022)Bakhtin, Wu, Lerer, Gray, Jacob, Farina, Miller,
  and Brown}]{bakhtin2022mastering}
Bakhtin, A.; Wu, D.~J.; Lerer, A.; Gray, J.; Jacob, A.~P.; Farina, G.; Miller,
  A.~H.; and Brown, N. 2022.
\newblock Mastering the Game of No-Press Diplomacy via Human-Regularized
  Reinforcement Learning and Planning.
\newblock \emph{arXiv preprint arXiv:2210.05492}.

\bibitem[{Barrett et~al.(2016)Barrett, Rosenfeld, Kraus, and
  Stone}]{AIJ16-Barrett}
Barrett, S.; Rosenfeld, A.; Kraus, S.; and Stone, P. 2016.
\newblock Making Friends on the Fly: Cooperating with New Teammates.
\newblock \emph{Artificial Intelligence}.

\bibitem[{Chakraborty and Stone(2014)}]{chakraborty2014convergence}
Chakraborty, D.; and Stone, P. 2014.
\newblock Convergence, targeted optimality and safety in multiagent learning.
\newblock \emph{Sample Efficient Multiagent Learning in the Presence of
  Markovian Agents}, 29--47.

\bibitem[{Charakorn, Manoonpong, and
  Dilokthanakul(2023)}]{charakorn2023generating}
Charakorn, R.; Manoonpong, P.; and Dilokthanakul, N. 2023.
\newblock Generating Diverse Cooperative Agents by Learning Incompatible
  Policies.
\newblock In \emph{The Eleventh International Conference on Learning
  Representations}.

\bibitem[{Christianos, Schäfer, and Albrecht(2020)}]{christianos2020shared}
Christianos, F.; Schäfer, L.; and Albrecht, S.~V. 2020.
\newblock Shared Experience Actor-Critic for Multi-Agent Reinforcement
  Learning.
\newblock In \emph{Advances in Neural Information Processing Systems
  (NeurIPS)}.

\bibitem[{Cui et~al.(2023{\natexlab{a}})Cui, Lupu, Sokota, Hu, Wu, and
  Foerster}]{cui2023adversarial}
Cui, B.; Lupu, A.; Sokota, S.; Hu, H.; Wu, D.~J.; and Foerster, J.~N.
  2023{\natexlab{a}}.
\newblock Adversarial Diversity in Hanabi.
\newblock In \emph{The Eleventh International Conference on Learning
  Representations}.

\bibitem[{Cui et~al.(2023{\natexlab{b}})Cui, Yang, Luo, Lee, Stone, Lee, Lee,
  Suh, Xiong, and Tian}]{cui2023macta}
Cui, J.; Yang, X.; Luo, M.; Lee, G.; Stone, P.; Lee, H.-H.~S.; Lee, B.; Suh,
  G.~E.; Xiong, W.; and Tian, Y. 2023{\natexlab{b}}.
\newblock {MACTA}: A Multi-agent Reinforcement Learning Approach for Cache
  Timing Attacks and Detection.
\newblock In \emph{The Eleventh International Conference on Learning
  Representations}.

\bibitem[{Duan et~al.(2016)Duan, Schulman, Chen, Bartlett, Sutskever, and
  Abbeel}]{duan2016rl}
Duan, Y.; Schulman, J.; Chen, X.; Bartlett, P.~L.; Sutskever, I.; and Abbeel,
  P. 2016.
\newblock $\text{RL}^{2}$: Fast reinforcement learning via slow reinforcement
  learning.

\bibitem[{Gu et~al.(2021)Gu, Zhao, Hao, and An}]{gu2021online}
Gu, P.; Zhao, M.; Hao, J.; and An, B. 2021.
\newblock Online ad hoc teamwork under partial observability.
\newblock In \emph{International Conference on Learning Representations}.

\bibitem[{Hu et~al.(2020)Hu, Lerer, Peysakhovich, and
  Foerster}]{hu2020otherplay}
Hu, H.; Lerer, A.; Peysakhovich, A.; and Foerster, J. 2020.
\newblock “other-play” for zero-shot coordination.
\newblock In \emph{International Conference on Machine Learning}, 4399--4410.
  PMLR.

\bibitem[{Lupu et~al.(2021)Lupu, Cui, Hu, and Foerster}]{lupu2021trajectory}
Lupu, A.; Cui, B.; Hu, H.; and Foerster, J. 2021.
\newblock Trajectory diversity for zero-shot coordination.
\newblock In \emph{International conference on machine learning}, 7204--7213.
  PMLR.

\bibitem[{Mahajan et~al.(2019)Mahajan, Rashid, Samvelyan, and
  Whiteson}]{mahajan2019maven}
Mahajan, A.; Rashid, T.; Samvelyan, M.; and Whiteson, S. 2019.
\newblock Maven: Multi-agent variational exploration.
\newblock \emph{Advances in Neural Information Processing Systems}, 32.

\bibitem[{Mirsky et~al.(2022)Mirsky, Carlucho, Rahman, Fosong, Macke,
  Sridharan, Stone, and Albrecht}]{mirsky2022survey}
Mirsky, R.; Carlucho, I.; Rahman, A.; Fosong, E.; Macke, W.; Sridharan, M.;
  Stone, P.; and Albrecht, S.~V. 2022.
\newblock A survey of ad hoc teamwork research.
\newblock In \emph{European Conference on Multi-Agent Systems}, 275--293.
  Springer.

\bibitem[{Papoudakis, Christianos, and Albrecht(2021)}]{papoudakis2021agent}
Papoudakis, G.; Christianos, F.; and Albrecht, S. 2021.
\newblock Agent Modelling under Partial Observability for Deep Reinforcement
  Learning.
\newblock \emph{Advances in Neural Information Processing Systems}, 35.

\bibitem[{Rahman et~al.(2023)Rahman, Fosong, Carlucho, and
  Albrecht}]{rahman2023generating}
Rahman, A.; Fosong, E.; Carlucho, I.; and Albrecht, S.~V. 2023.
\newblock Generating Teammates for Training Robust Ad Hoc Teamwork Agents via
  Best-Response Diversity.
\newblock \emph{Transactions on Machine Learning Research}.

\bibitem[{Rahman et~al.(2021)Rahman, H{\"o}pner, Christianos, and
  Albrecht}]{rahmanOpenAdHoc2021}
Rahman, A.; H{\"o}pner, N.; Christianos, F.; and Albrecht, S.~V. 2021.
\newblock Towards Open Ad Hoc Teamwork Using Graph-Based Policy Learning.
\newblock In \emph{{{International Conference}} on {{Machine Learning}}},
  volume 139. {PMLR}.

\bibitem[{Stone et~al.(2010)Stone, Kaminka, Kraus, and
  Rosenschein}]{stone2010adhoc}
Stone, P.; Kaminka, G.; Kraus, S.; and Rosenschein, J. 2010.
\newblock Ad hoc autonomous agent teams: Collaboration without
  pre-coordination.
\newblock In \emph{Proceedings of the AAAI Conference on Artificial
  Intelligence}, volume~24, 1504--1509.

\bibitem[{Strouse et~al.(2021)Strouse, McKee, Botvinick, Hughes, and
  Everett}]{strouse2021fititiouscoplay}
Strouse, D.; McKee, K.; Botvinick, M.; Hughes, E.; and Everett, R. 2021.
\newblock Collaborating with humans without human data.
\newblock \emph{Advances in Neural Information Processing Systems}, 34:
  14502--14515.

\bibitem[{Vinyals et~al.(2019)Vinyals, Babuschkin, Czarnecki, Mathieu, Dudzik,
  Chung, Choi, Powell, Ewalds, Georgiev et~al.}]{vinyals2019grandmaster}
Vinyals, O.; Babuschkin, I.; Czarnecki, W.~M.; Mathieu, M.; Dudzik, A.; Chung,
  J.; Choi, D.~H.; Powell, R.; Ewalds, T.; Georgiev, P.; et~al. 2019.
\newblock Grandmaster level in StarCraft II using multi-agent reinforcement
  learning.
\newblock \emph{Nature}, 575(7782): 350--354.

\bibitem[{Xing et~al.(2021)Xing, Liu, Zheng, Pan, and Zhou}]{xing2021entropy}
Xing, D.; Liu, Q.; Zheng, Q.; Pan, G.; and Zhou, Z. 2021.
\newblock Learning with Generated Teammates to Achieve Type-Free Ad-Hoc
  Teamwork.
\newblock In \emph{IJCAI}, 472--478.

\bibitem[{Yu et~al.(2022)Yu, Velu, Vinitsky, Gao, Wang, Bayen, and
  Wu}]{yu2022the}
Yu, C.; Velu, A.; Vinitsky, E.; Gao, J.; Wang, Y.; Bayen, A.; and Wu, Y. 2022.
\newblock The Surprising Effectiveness of {PPO} in Cooperative Multi-Agent
  Games.
\newblock In \emph{Thirty-sixth Conference on Neural Information Processing
  Systems Datasets and Benchmarks Track}.

\bibitem[{Zintgraf et~al.(2021)Zintgraf, Devlin, Ciosek, Whiteson, and
  Hofmann}]{zintgraf2021deep}
Zintgraf, L.; Devlin, S.; Ciosek, K.; Whiteson, S.; and Hofmann, K. 2021.
\newblock Deep interactive bayesian reinforcement learning via meta-learning.
\newblock \emph{arXiv preprint arXiv:2101.03864}.

\end{thebibliography}
\newpage
\flushbottom
\appendix
\section{Teammate Policies for AHT Evaluation}
\label{Appendix:EvalTeammateTypes}
We outline the different types of teammate policies in the set of teammates we use for AHT evaluation, $\Pi^{\text{eval}}$. For each environment, teammate policies in $\Pi^{\text{eval}}$ are based on simple heuristics. Details of heuristics used for each environment are outlined in the following sections.

\subsection{Repeated Matrix Game}
\label{Apdx:RepeatedMatGame}
Since the Repeated Matrix Game is a simple environment without any states, we only implemented six simple heuristics which details are provided below:
\begin{itemize}
    \item \textbf{H1}. Agents that follow this heuristic will always choose the first action.
    \item \textbf{H2}. This heuristic will get an agent to always choose the second action.
    \item \textbf{H3}. Agents using this heuristic will always choose the third action.
    \item \textbf{H4}. Unlike H1-H3, this heuristic gives agents a policy that chooses the first, second, and third action with probabilities of 0.7, 0.15, and 0.15 respectively.
    \item \textbf{H5}. This is a policy that chooses the first, second, and third action with probabilities of 0.15, 0.7, and 0.15 respectively.
    \item \textbf{H6}. Agents following this heuristic will choose the third action 70\% of the time. Meanwhile, it is also equally likely to choose between the first and second actions.

\end{itemize}

\subsection{Cooperative Reaching and Weighted Cooperative Reaching}
For the Cooperative Reaching and Weighted Cooperative Reaching environment, we implement 15 types of teammate heuristics whose behaviour are detailed below:
\begin{itemize}
    \item \textbf{H1.} H1 controls an agent to always move to the closest corner cell from its initial location.
    \item \textbf{H2.} This heuristic moves an agent towards the furthest corner cell from its the agent's initial location at the beginning of the episode.
    \item \textbf{H3.} H3 controls an agent to move towards the closest corner cell between corner cells A and B.
    \item \textbf{H4.} Based on the agent's initial location at the beginning of an episode, H4 will move agents towards the furthest cell between cells A and B.
    \item \textbf{H5.} H5 moves an agent towards the closest cell between cells C and D.
    \item \textbf{H6.} Depending on the agent's position at the beginning of an episode, H6 controls the agent to move towards the furthest cell between cells C and D.
    \item \textbf{H7.} At the beginning of each interaction, H7 randomly picks a destination cell between A, B, C, and D with equal probability. For the remainder of each episode, the agent will be controlled to move towards the destination cell.
    \item \textbf{H8-H11}. H8-H11 move agents towards corner cells A-D respectively.
    \item \textbf{H12}. H12 moves an agent towards corner cell A with a 55\% chance. Meanwhile, the other corner cells are equally likely to be chosen as destination cells.
    \item \textbf{H13}. H13 moves an agent towards corner cells A, B, C, and D with a 15\%, 55\%, 15\%, and 15\% chance respectively. 
    \item \textbf{H14}. H14 moves an agent towards corner cells A, B, C, and D with a 15\%, 15\%, 55\%, and 15\% chance respectively. 
    \item \textbf{H15}. H15 moves an agent towards corner cell D 55\% of the time. Meanwhile, the remaining corner cells are equally likely to be chosen as destination cells.
\end{itemize}

\subsection{Level-based Foraging}
Experiments in the Level-based Foraging environment evaluate AHT agents against $\Pi^{\text{eval}}$ consisting of 8 heuristic types defined below:
\begin{itemize}
    \item \textbf{H1.} Agents under H1 will move towards the closest item from its current location and collect it. This process is repeated until no item is left.
    \item \textbf{H2.} At the beginning of an episode, agents under heuristic H2 will move towards the furthest object from its location and collect it. Every time its targeted item is collected, the agent will then move to collect the remaining item whose location is furthest from the agent's current location. This process is repeated until no item remains.
    \item \textbf{H3-H8.} H3-H8 each corresponds to a heuristic that collects items following one of the six possible permutations of collecting the three items available in the environment.
\end{itemize}

\section{Analyzing Baseline Failure in Repeated Matrix Game \& Weighted Cooperative Reaching}
\label{Appendix:BaselineFailure}
In this section, we mathematically demonstrate that no constant and uniform $\alpha > 0$ can make BRDiv or LIPO identify all policies in MCS(E) for the Repeated Matrix Game and Weighted Cooperative Reaching environment. Section~\ref{Apdx:RepeatedMatrixGame} details our argument regarding the baselines' failure in the repeated matrix game. Meanwhile, the same argument for the Weighted Cooperative Reaching environment is provided in Section~\ref{Apdx:WeightedCoopReaching}.

\subsection{Repeated Matrix Game}
\label{Apdx:RepeatedMatrixGame}

Based on the payoff matrix provided in Figure~\ref{Fig:RPMEnv}, it is clear that the MCS of the Repeated Matrix Game environment consists of the three deterministic policies displayed in Figure~\ref{Fig:MatrixActionFrequency}. Ideally, L-BRDiv, BRDiv, and LIPO should all produce MCS${}^{\text{est}}(\text{E})$ and $\Pi^{\text{train}}$ containing policies displayed in Figure~\ref{Fig:MatrixActionFrequency}. However, we show it is impossible to find $\alpha > 0$ that can make BRDiv and LIPO discover MCS${}^{\text{est}}(\text{E})$ for this environment and generate a set of teammate policies to maximize the AHT agent's robustness.

LIPO and BRDiv fail in this simple environment because another set of policies produces a higher adversarial diversity metric compared to the ideal MCS${}^{\text{est}}(\text{E})$ and $\Pi^{\text{train}}$ for any $\alpha > 0$. An example set of policies producing a higher adversarial diversity metric than the ideal MCS${}^{\text{est}}(\text{E})$ is displayed in Figure~\ref{Fig:MatrixGameFailure}. Compared to discovering MCS(E) as MCS${}^{\text{est}}(\text{E})$ and $\Pi^{\text{train}}$ that results in a cross-play matrix like the payoff matrix, the cross-play matrix from discovering policies in Figure~\ref{Fig:MatrixActionFrequencyAlternative} has a lower sum of non-diagonal elements while having the same trace. 

\begin{figure}[t]
\begin{minipage}[h]{\linewidth}
\begin{center}
\begin{adjustbox}{width=0.5\textwidth}
\begin{tabular}{|c|c|c|c|}
\hline
  & $\pi$(A)  & $\pi$(B)  & $\pi$(C)  \\ \hline
1 & 1 & 0  & 0  \\ \hline
2 & 0  & 1 & 0  \\ \hline
3 & 0  & 1  & 0 \\ \hline
\end{tabular}
\end{adjustbox}
\end{center} 
\subcaption{A set of policies that appear more optimal than MCS(E) for BRDiv and LIPO.}
\label{Fig:MatrixActionFrequencyAlternative}
\end{minipage}

\begin{minipage}[h]{\linewidth}
\begin{center}
\begin{adjustbox}{width=0.3\textwidth}
\begin{tabular}{|c|c|c|}
\hline
10 & 0  & 0  \\ \hline
0  & 6 & 6  \\ \hline
0  & 6  & 6 \\ \hline
\end{tabular}
\end{adjustbox}
\label{Fig:ComparedXPMatrix}
\end{center} 
\subcaption{Cross-play matrix for the policies discovered in Figure~\ref{Fig:MatrixActionFrequencyAlternative}.}
\end{minipage}
\caption{An Example Failure Mode of BRDiv \& LIPO. The above figures provide an example set of policies that will appear to be more optimal than MCS(E) if we optimize the diversity metric used by LIPO and BRDiv.}
\label{Fig:MatrixGameFailure}
\end{figure}

We now evaluate the value of LIPO and BRDiv's optimized diversity metric when  both MCS${}^{\text{est}}(\text{E})$ and $\Pi^{\text{train}}$ equals MCS(E) and when it instead discovers the set of policies displayed in Figure~\ref{Fig:MatrixActionFrequencyAlternative}, which we denote as $\Pi^{\text{alt}}$. Note that the adversarial diversity metric maximized by BRDiv, BRDiv($\{\pi^{i}\}_{i=1}^{K}$,$\{\pi^{-i}\}_{i=1}^{K}$), can be expressed as:
\begin{align}
    &\sum_{i\in\{1,\dots,K\}} \mathbb{E}_{s_{0}\sim{p_{0}}}\left[\textbf{R}_{{i},{-i}}(H_{t})\right] + \nonumber\\ &\sum_{\substack{i,j\in\{1,\dots,K\}\\i\neq j}} \text{ }\alpha \left(  \mathbb{E}_{s_{0}\sim{p_{0}}}\left[\textbf{R}_{i,-i}(H_{t}) - \textbf{R}_{{j},{-i}}(H_{t})\right] \right) + \nonumber\\
    &\sum_{\substack{i,j\in\{1,\dots,K\}\\i\neq j}} \text{ }\alpha\left(\mathbb{E}_{s_{0}\sim{p_{0}}}\left[\textbf{R}_{{i},{-i}}(H_{t}) - \textbf{R}_{{i},{-j}}(H_{t})\right]\right),
\end{align}
for some $\alpha > 0$. Meanwhile, the adversarial diversity metric optimized by LIPO, LIPO($\{\pi^{i}\}_{i=1}^{K}$,$\{\pi^{-i}\}_{i=1}^{K}$), is given by the following expression:
    \begin{align}
    &\sum_{i\in\{1,\dots,K\}} \mathbb{E}_{s_{0}\sim{p_{0}}}\left[\textbf{R}_{{i},{-i}}(H_{t})\right] - \nonumber\\ &\sum_{\substack{i,j\in\{1,\dots,K\}\\i\neq j}} \text{ }\alpha \left(  \mathbb{E}_{s_{0}\sim{p_{0}}}\left[ \textbf{R}_{{j},{-i}}(H_{t}) + \textbf{R}_{{i},{-j}}(H_{t})\right] \right),
    \label{Eq:LIPOOpt}
\end{align}
assuming $\alpha > 0$. For $\alpha > 0$, the resulting BRDiv and LIPO objective for both sets of policies are provided in the following table:
\begin{table}[h]
\centering
\caption[Value of LIPO and BRDiv objectives for the Repeated Matrix Game.]{\textbf{Value of LIPO and BRDiv objectives for the Repeated Matrix Game.} The expressions that evaluate LIPO and BRDiv's optimized diversity metric for the Repeated Matrix Game are provided below. No $\alpha > 0$ enables MCS(E) to have higher diversity objectives than $\Pi^{\text{alt}}$.}
\label{Tab:LIPOBRDivObjMatrix}
\begin{tabular}{|c|c|c|}
\hline
Method  & \begin{tabular}[c]{@{}c@{}}MCS(E) \end{tabular} & $\Pi^{\text{alt}}$ \\ \hline
BRDiv   &  22+56$\alpha$                                                             & 22+64$\alpha$                                                                                             \\ \hline
LIPO    &  22-16$\alpha$                                                             & 22-12$\alpha$                                                                                             \\ \hline
\end{tabular}
\end{table}
From Table~\ref{Tab:LIPOBRDivObjMatrix}, it is clear that discovering $\Pi^{\text{alt}}$ will always produce higher diversity metrics for BRDiv and LIPO. It is then impossible to discover MCS(E) while optimizing both of these objectives. Its inability to discover some members of MCS(E) and instead discover other members twice eventually leads LIPO and BRDiv towards producing AHT agents with significantly worse returns than L-BRDiv.

\subsection{Weighted Cooperative Reaching}
\label{Apdx:WeightedCoopReaching}

\begin{figure}[t]
\begin{minipage}[h]{\linewidth}
\begin{center}
\begin{adjustbox}{width=0.65\textwidth}
\begin{tabular}{|c|c|c|c|c|}
\hline
  & $\pi$(A)  & $\pi$(B)  & $\pi$(C)& $\pi$(D)  \\ \hline
1 & 1 & 0  & 0 & 0  \\ \hline
2 & 1  & 0 & 0 & 0  \\ \hline
3 & 0  & 1  & 0 & 0\\ \hline
4 & 0  & 1  & 0 & 0\\ \hline
\end{tabular}
\end{adjustbox}
\end{center} 
\subcaption{Denoting $\pi(\text{X})$ as the probability of ending up in a corner cell having an ID of X, the above set of policies produce higher diversity metrics than MCS(E) in the Weighted Cooperative Reaching environment for BRDiv and LIPO.}
\label{Fig:WeightCoopPoliciesAlternative}
\end{minipage}

\begin{minipage}[h]{\linewidth}
\begin{center}
\begin{adjustbox}{width=0.45\textwidth}
\begin{tabular}{|c|c|c|c|}
\hline
10 & 10 & 0  & 0  \\ \hline
10  & 10 & 0 & 0 \\ \hline
0  & 0  & 10 & 10 \\ \hline
0  & 0  & 10 & 10 \\ \hline
\end{tabular}
\end{adjustbox}
\end{center} 
\subcaption{Cross-play matrix between policies discovered in Figure~\ref{Fig:WeightCoopPoliciesAlternative}.}
\label{Fig:ComparedXPMatrixWeighCoop}
\end{minipage}
\caption{Another Example Failure Mode of BRDiv \& LIPO in Weighted Cooperative Reaching. By not discovering policies that move towards corner cells C and D, BRDiv and LIPO can achieve a higher diversity metric than when discovering MCS(E).}
\label{Fig:WeighCoopFailure}
\end{figure}

To show the shortcomings of LIPO and BRDiv in Weighted Cooperative Reaching, we also construct a set of policies that will produce higher diversity metrics for both BRDiv and LIPO. This set of policies that appears more desirable for LIPO and BRDiv than MCS(E) is denoted by $\Pi^{\text{alt}}$ and is visualized by Figure~\ref{Fig:WeighCoopFailure}. Instead of discovering four policies moving towards different corner cells in the environment, $\Pi^{\text{alt}}$ discovers policies moving towards cells A and B twice. Discovering $\Pi^{\text{alt}}$ and using it as MCS${}^{\text{est}}(\text{E})$ and $\Pi^{\text{train}}$ results in a cross-play matrix displayed in Figure~\ref{Fig:ComparedXPMatrixWeighCoop}.

Compared to MCS(E) that produces a cross-play matrix that is the same as Figure~\ref{Fig:WeightedCoopReachingEnv}, the cross-play matrix from $\Pi^{\text{alt}}$ has a higher sum of self-play returns and a lower sum of cross-play returns. As a result, no $\alpha > 0$ should make MCS(E) appear more desirable to LIPO and BRDiv. We show the expressions evaluating LIPO and BRDiv's diversity metrics for MCS(E) and $\Pi^{\text{alt}}$ in Table~\ref{Tab:LIPOBRDivObjMatrixWeigh}. Since a set of policies like $\Pi^{\text{alt}}$ that does not discover all members of MCS(E) appear more preferable than MCS(E), LIPO and BRDiv end up yielding AHT agents that cannot robustly interact with teammate policies whose best-response policies are not discovered. 

\begin{table}[h]
\centering
\caption[Value of LIPO and BRDiv objectives for Weighted Cooperative Reaching.]{Value of LIPO \& BRDiv Objectives for Weighted Cooperative Reaching. The expressions that evaluate LIPO and BRDiv's optimized diversity metric for Weighted Cooperative Reaching are provided below. No $\alpha > 0$ enables MCS(E) to have higher diversity objectives than $\Pi^{\text{alt}}$.}
\label{Tab:LIPOBRDivObjMatrixWeigh}
\begin{tabular}{|c|c|c|}
\hline
Method  & \begin{tabular}[c]{@{}c@{}}MCS(E) \end{tabular} & $\Pi^{\text{alt}}$ \\ \hline
BRDiv   &  36+120$\alpha$                                                             & 40+160$\alpha$                                                                                             \\ \hline
LIPO    &  36-48$\alpha$                                                             & 40-40$\alpha$                                                                                             \\ \hline
\end{tabular}
\end{table}

\section{Analyzing the Lagrange Multipliers of L-BRDiv}
\label{sec:LagrangeAnalysis}
\begin{figure*}[t]
\includegraphics{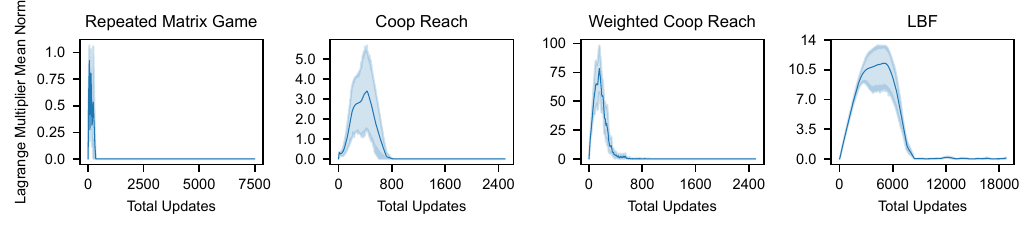}

\caption{The Changing Values of L-BRDiv's Lagrange Multipliers. Figure~\ref{Figs:LagrangeChange} show how L-BRDiv's Lagrange multipliers change over time. Since a randomly initialized policy will not fulfil the constraints upheld by L-BRDiv, the Lagrange multipliers will initially increase their value to add more pressure to the policies to fulfil the constraints. Finally, the Lagrange multipliers will decrease to zero once constraints are fulfilled.} 
\label{Figs:LagrangeChange}
\end{figure*}

The role of the Lagrange multipliers in the learning process undergone by L-BRDiv is highlighted in Figure~\ref{Figs:LagrangeChange}. Since the randomly initialized teammate policies cannot fulfil the upheld constraints in the beginning, optimizing Expression~\ref{def:dual_lagrange} encourages the increase of the Lagrange multipliers' values. The increasingly large Lagrange multipliers then force the learned policies to start fulfilling these constraints. Once policies learn to fulfil a constraint, the Lagrange multiplier associated with that constraint will decrease towards zero. At the end of the optimization process, we see that all Lagrange multipliers eventually converge to zero after all constraints are fulfilled.

\section{Computational Complexity of L-BRDiv}
\label{sec:CompComplexity}

The complexity of a neural network's forward computation and backpropagation will then serve as a basis to identify the computational complexity of L-BRDiv. Denoting the size of the n${}^{th}$ hidden layer of the policy network as $L_{n}$ and given input data with |$D$| datapoints \& $F$ features, the computational complexity of forward computation and stochastic gradient descent (SGD) for neural networks is $\mathcal{O}$($|D|M)$ with $M=\text{max}(FL_{1}, \text{max}_{i}(L_{i}L_{i+1}))$. This complexity follows from forward and backpropagation in neural networks being a sequence of matrix multiplications. 

Given MAPPO, BRDiv, and LIPO's experience collection and policy update process (based on optimizing Expressions~\ref{def:dual_lagrange} and~\ref{Eq:WeightedAdvantage} for a given $\alpha$ described in Appendix~\ref{sec:LagrangeAnalysis}) that does forward and backpropagation for all $T$ experiences collected during training, their complexity becomes $\mathcal{O}$($TM)$. Unlike these methods, L-BRDiv also has to compute the Lagrange dual for each experience (Line 12 in pseudocode). Given $K$ generated policies, computing the Lagrange dual for each requires computing $2K(K-1)$ forward computations for each experience, which results in a $\mathcal{O}$($K^{2}TM)$ complexity. Although it may appear to be a considerable increase, note that $K$ is often set to a small value. Existing neural network libraries can also parallelize the $2K(K-1)$ forward computations in the Lagrange dual evaluation using GPUs, resulting in a computational complexity closer to $\mathcal{O}$($TM)$ for L-BRDiv. 

\section{Teammate Generation Hyperparameter Details}
\label{Appdx:Hyperparam}
The hyperparameters that we use during L-BRDiv's teammate generation process are provided below:
\begin{itemize}
    \item $K$: Number of generated policies.
    \item $\lambda_{\pi}$: Policy learning rate.
    \item $\lambda_{V}$: Critic learning rate.
    \item $\lambda_{\alpha}$: Lagrange multiplier learning rate.
    \item $\gamma$: Discount rate.
    \item $T$: Number of experiences used in learning.
    \item $N_{\text{threads}}$: Number of parallel threads for data collection during training.
    \item $T_{\text{update}}$: Number of timesteps between update.
    \item $T_{\text{lagrange}}$: Number of policy updates between subsequent Lagrange multiplier updates.
    \item $\tau$: Tolerance factor used in the Lagrange dual.
    \item $w_{\text{ent}}$: Entropy multiplier to encourage exploration in MAPPO. To prevent the magnitude of the entropy loss from being overwhelmed by the policy loss, in practice we multiply this term with $w^{i,i}(\mathbb{A})$ in Expression~\ref{Eq:WeightedAdvantage} to compute the entropy weights.
\end{itemize}
For these hyperparameters, we outline their value for the four environments used in our experiments as provided in Table~\ref{Tab:L-BRDivHyperParams}. Meanwhile, we also use multilayer perceptrons as our policy and critic network architecture for all compared methods. Details of the size of these models in each environment are provided in Table~\ref{Tab:L-BRDivNetSize}.
\begin{table}[!t]
\centering
\caption[Hyperparameter values for L-BRDiv's Experiments.]{Hyperparameter Values for L-BRDiv's Experiments. The specific hyperparameter values used in our teammate generation experiments in Repeated Matrix Games (RPM), Cooperative Reaching (CR), Weighted Cooperative Reaching (WCR), and Level-based Foraging (LBF) are provided below.}
\label{Tab:L-BRDivHyperParams}
\begin{tabular}{|c|c|c|c|c|}
\hline
                      & RPM       & CR                                                & WCR                                               & LBF                \\ \hline
$K$                   & 3         & 4                                                             & 4                                                             & 6                  \\ \hline
$\lambda_{\pi}$       & $10^{-3}$ & $10^{-4}$                                                     & $10^{-4}$                                                     & $10^{-4}$          \\ \hline
$\lambda_{V}$         & $10^{-3}$ & $10^{-4}$                                                     & $10^{-4}$                                                     & $10^{-4}$          \\ \hline
$\lambda_{\alpha}$    & 0.05      & 0.5                                                           & 0.5                                                           & 0.05               \\ \hline
$\gamma$              & 0.99      & 0.99                                                          & 0.99                                                          & 0.99               \\ \hline
$T$                   & $10^{6}$  & \begin{tabular}[c]{@{}c@{}}3.2$\times10^{7}$\end{tabular} & \begin{tabular}[c]{@{}c@{}}3.2$\times10^{7}$\end{tabular} & 2.4$\times 10^{8}$ \\ \hline
$N_{\text{threads}}$  & 40        & 160                                                           & 160                                                           & 160                \\ \hline
$T_{\text{update}}$   & 2         & 8                                                             & 8                                                             & 8                  \\ \hline
$T_{\text{lagrange}}$ & 10        & 10                                                            & 10                                                            & 10                 \\ \hline
$\tau$                & 1         & 0.2                                                           & 0.5                                                           & 0.1                \\ \hline
$w_{\text{ent}}$                & $10^{-3}$         & \begin{tabular}[c]{@{}c@{}}5$\times10^{-3}$\end{tabular}                                                       & 5$\times 10^{-3}$                                                           & 8$\times 10^{-4}$                \\ \hline
\end{tabular}
\end{table}

\begin{table}[t]
\centering
\caption[Network size for L-BRDiv's experiments.]{Network size for L-BRDiv's experiments. The size of models in our experiments in the Repeated Matrix Games (RPM), Cooperative Reaching (CR), Weighted Cooperative Reaching (WCR), and Level-based Foraging (LBF) environment are detailed below.}
\label{Tab:L-BRDivNetSize}
\begin{tabular}{|c|c|c|c|c|}
\hline
                      & RPM       & CR                                                & WCR                                               & LBF                \\ \hline
$\pi^{i}_{\theta}$ (Layer 1)                  & 32        & 128                                                             & 128                                                             & 128                   \\ \hline
$\pi^{i}_{\theta}$ (Layer 2)       & 32 & 256                                                     & 256                                                     & 128          \\ \hline
$\pi^{i}_{\theta}$ (Layer 3)       & N/A & 256                                                     & 256                                                     & N/A          \\ \hline
$\pi^{i}_{\theta}$ (Layer 4)       & N/A & 128                                                     & 128                                                     & N/A          \\ \hline
$V_{\theta_{c}}$ (Layer 1)         & 32 & 128                                                     & 128                                                     & 128          \\ \hline
$V_{\theta_{c}}$ (Layer 2)    & 32      & 256                                                          & 256                                                           & 128               \\ \hline
$V_{\theta_{c}}$ (Layer 3)    & N/A      & 256                                                           & 256                                                           & N/A               \\ \hline
$V_{\theta_{c}}$ (Layer 4)    & N/A      & 128                                                          & 128                                                           & N/A               \\ \hline
\end{tabular}
\end{table}

We ensure a fair comparison between L-BRDiv and the baseline methods by using the same hyperparameter values and network architecture. However, note that BRDiv and LIPO still require us to set $\alpha$ to a value that facilitates the generation of $\Pi^{\text{train}}$ that facilitates the training of robust AHT agents. Since teammate generation and AHT training is computationally expensive , we follow these steps to tune $\alpha$:
\begin{enumerate}
    \item We initially run LIPO and BRDiv with $\alpha\in\{0.1, 0.5, 1, 5, 10\}$. Two experiment runs are done for each $\alpha$.
    \item We look at the generated teammates and see which tested $\alpha$ discover more members of MCS(E). 
    \item Based on the $\alpha$ producing the best estimate of MCS(E), we then do slight tuning to $\alpha$ by finding values close to $\alpha$ producing the best approximate to MCS(E).
\end{enumerate}
Following this process, the final hyperparameter value that we end up using for LIPO and BRDiv is summarized in Table~\ref{Tab:BaselineAlpha}. In alignment with the findings from~\citet{charakorn2023generating}, note that LIPO ends up using small $\alpha$ values since larger $\alpha$ results in incompetent policies that cannot even achieve high returns against their intended partner in self-play. The only exception is Cooperative Reaching where MCS(E) consists of policies whose cross-play returns are zero, which enables the use of a large $\alpha$. This emergence of incompetent policies is a natural consequence of optimizing Expression~\ref{Eq:LIPOOpt}, which cross-play return term's magnitude can overwhelm the self-play return term for large enough $\alpha$.

\begin{table}[h]
\centering
\caption[Weight hyperparameter for baseline methods.]{$\mathbf{\alpha}$ for Baseline Methods. The value of $\alpha$ used by baseline methods in their respective objectives for the Repeated Matrix Games (RPM), Cooperative Reaching (CR), Weighted Cooperative Reaching (WCR), and Level-based Foraging (LBF) environment are detailed below.}
\label{Tab:BaselineAlpha}
\begin{tabular}{|c|c|c|c|c|}
\hline
                      & RPM       & CR                                                & WCR                                               & LBF                \\ \hline
LIPO                 & 0.5        & 8                                                             & 0.25                                                             & 0.08                   \\ \hline
BRDiv                 & 1       & 10                                                             & 1                                                             & 0.4                   \\ \hline
\end{tabular}
\end{table}

\section{AHT Experiment Hyperparameters}
As we mention in Section~\ref{sec:ExperimentSetup}, we use the RL${}^{2}$ algorithm to train AHT agents based on the set of teammates generated by each compared method. The hyperparameters of the RL${}^{2}$ algorithm are listed below:
\begin{itemize}
    \item $\lambda_{\pi}$: Policy learning rate.
    \item $\lambda_{V}$: Critic learning rate.
    \item $\gamma$: Discount rate.
    \item $T$: Number of experiences used in learning.
    \item $N_{\text{threads}}$: Number of parallel threads for data collection during training.
    \item $T_{\text{update}}$: Number of timesteps between update.
    \item $w_{\text{ent}}$: Entropy weight term to encourage exploration.
    \item $L_{\text{rep}}$: The length of representation vectors to characterize teammates.
\end{itemize}
For each environment used in our experiments, hyperparameter values that we use in each environment is provided in Table~\ref{Tab:AHTExpHyperParams}. 

\begin{table}[!h]
\centering

\caption[Hyperparameter values for AHT Experiments.]{Hyperparameter values for L-BRDiv's Experiments. The specific hyperparameter values used in our Repeated Matrix Games (RPM), Cooperative Reaching (CR), Weighted Cooperative Reaching (WCR), and Level-based Foraging (LBF) environment are provided below.}
\label{Tab:AHTExpHyperParams}
\begin{tabular}{|c|c|c|c|c|}
\hline
                      & RPM       & CR                                                & WCR                                               & LBF                \\ \hline
$\lambda_{\pi}$       & $10^{-4}$ & $10^{-4}$                                                     & $10^{-4}$                                                     & $10^{-4}$          \\ \hline
$\lambda_{V}$         & $10^{-4}$ & $10^{-4}$                                                     & $10^{-4}$                                                     & $10^{-4}$          \\ \hline
$\gamma$              & 0.99      & 0.99                                                          & 0.99                                                          & 0.99               \\ \hline
$T$                   & $10^{6}$  & \begin{tabular}[c]{@{}c@{}}1.2$\times10^{7}$\end{tabular} & \begin{tabular}[c]{@{}c@{}}1.2$\times10^{7}$\end{tabular} & 4.8$\times 10^{7}$ \\ \hline
$N_{\text{threads}}$  & 10        & 16                                                           & 16                                                           & 16                \\ \hline
$T_{\text{update}}$   & 2         & 8                                                             & 8                                                             & 8                  \\ \hline
$w_{\text{ent}}$                & $10^{-4}$         & \begin{tabular}[c]{@{}c@{}}2.5$\times10^{-4}$\end{tabular}                                                       & 2.5$\times 10^{-4}$                                                           & 8$\times 10^{-4}$                \\ \hline
$L_{\text{rep}}$                & 16         & 32                                                       & 32                                                           & 64                \\ \hline
\end{tabular}
\end{table}

Apart from these hyperparameters, our policy and critic networks have a similar architecture to the teammate generation process. The only difference is that we use an LSTM layer as our final layer. We use the LSTM layer to enable agents to process the previous sequence of observations and experienced rewards to model the type of teammates the AHT agent is interacting with.

\end{document}